\documentclass[10pt, a4paper]{article}

\usepackage[table,xcdraw]{xcolor}

\usepackage[final]{lrec2026} 

\usepackage{booktabs}
\usepackage{multirow}

\definecolor{lightpurple}{RGB}{166,172,213}
\definecolor{lightblue}{RGB}{166,213,213}
\definecolor{lightpink}{RGB}{213,166,189}

\title{RespondeoQA: a Benchmark for Bilingual Latin-English Question Answering}

\name{Marisa Hudspeth\textsuperscript{1} \quad
        Patrick J. Burns\textsuperscript{2} \quad
        Brendan O'Connor\textsuperscript{1}} 

\address{\textsuperscript{1}Manning College of Information \& Computer Sciences, University of Massachusetts Amherst \\
  \textsuperscript{2}Institute for the Study of the Ancient World, New York University \\
         \{mhudspeth,brenocon\}@cs.umass.edu \quad
            pjb311@nyu.edu
         }

\abstract{
We introduce a benchmark dataset for question answering and translation in bilingual Latin and English settings, containing about 7,800 question–answer pairs. The questions are drawn from Latin pedagogical sources, including exams, quizbowl-style trivia, and textbooks ranging from the 1800s to the present. After automated extraction, cleaning, and manual review, the dataset covers a diverse range of question types: knowledge- and skill-based, multihop reasoning, constrained translation, and mixed language pairs. To our knowledge, this is the first QA benchmark centered on Latin. As a case study, we evaluate three large language models--LLaMa 3, Qwen QwQ, and OpenAI's o3-mini--finding that all perform worse on skill-oriented questions. Although the reasoning models perform better on scansion and literary-device tasks, they offer limited improvement overall. QwQ performs slightly better on questions asked in Latin, but LLaMa3 and o3-mini are more task dependent.
This dataset provides a new resource for assessing model capabilities in a specialized linguistic and cultural domain, and the creation process can be easily adapted for other languages.
The dataset is available at: \href{https://github.com/slanglab/RespondeoQA}{https://github.com/slanglab/RespondeoQA}
 \\ \newline \Keywords{historical languages, evaluation, question answering} }

\begin{document}

\maketitleabstract


\section{Introduction}

In recent years, large language models (LLMs) have shown impressive abilities across a wide range of natural language understanding and generation tasks. 
Yet their performance on many languages, including historical ones like Latin,
remains underexplored. Latin occupies a unique position compared to other languages: it is no longer spoken, but has a rich written tradition spanning over two millennia and remains a cornerstone of classical education \cite{leonhardt2013latin}. Because of this history, Latin is a highly frequent language in large-scale archival sources---for example, it is the 5th most prevalent language in two recently released corpora, Common Corpus \cite{langlais_2025} and Institutional Books \cite{cargnelutti_2025}.

Despite the abundance of Latin textual data, few resources exist for evaluating generative LLMs' capabilities for Latin cultural and language skills. Most Latin-specific datasets are designed for token or sentence level classification tasks more suitable for encoder models (NER, WSD, POS tagging, others), although recent work has begun exploring the abilities of generative LLMs for Latin \cite{gorovaia-etal-2024-sui, volk-etal-2024-llm, Marmonier2025ExplicitLA}. For machine translation specifically, there are few existing sentence-aligned datasets that are large enough for training or robust evaluation \cite{martinez-garcia-garcia-tejedor-2020-latin, fischer-etal-2022-machine, rosenthal2023machina}, so automatic sentence alignment methods are an open area of research. 
Often, automatic sentence alignment for Latin is performed from a digital humanities or corpus analysis perspective, focusing on existing, canonical sources that have been extensively translated and studied \cite{yousef-etal-2022-automatic-translation, Craig2023TestingTL}. These sentences and their translations are likely to have appeared frequently in LLM pretraining data, raising concerns that LLMs may reproduce memorized translations. Thus, these sentence-aligned datasets may be better suited for training rather than evaluation of LLMs.

\begin{table}[!t]
    \centering
    \small
    \setlength{\tabcolsep}{5pt}
    \begin{tabular}{p{0.46\linewidth} p{0.21\linewidth} p{0.20\linewidth}}
        \toprule
        \textbf{Data Source} & \textbf{Year(s)} & \textbf{Type} \\
        \midrule
        \textit{Exercises in Latin Prosody and Versification} & 1823 & OCR book scan \\ \hline
        \textit{Latin Grammar and Junior Scholarship Papers} & 1832 & OCR book scan \\ \hline
        Certamen & 1996--2009 & Digital text (MS Word) \\ \hline
        National Latin Exam (NLE) & 2015, 2020, 2025 & Digital text (PDF) \\
        \bottomrule
    \end{tabular}
    \caption{Sources of our QA data with year of publication and format type. For Certamen and NLE, we only list the years from which we obtained questions, but both have published materials in other years.}
    \label{tab:data_sources}
    \vspace{-2mm}
\end{table}


\begin{figure}[t]
    \centering

    \includegraphics[width=\linewidth]{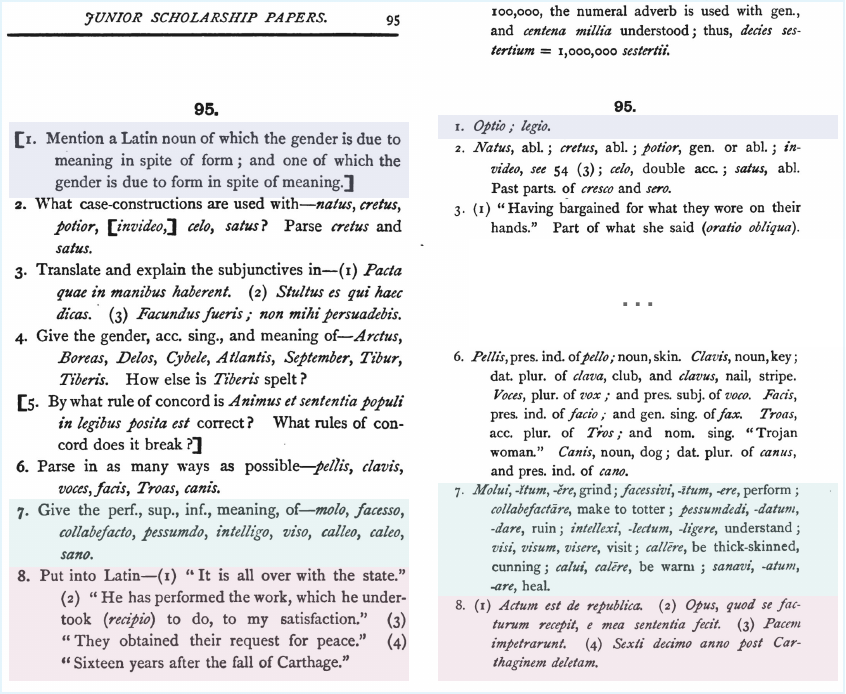}

    \caption{A page from \textit{Latin Grammar and Junior Scholarship Papers} (left) and its answer key (right).}
    \label{fig:pdf_ex}
    \vspace{-4mm}
\end{figure}

\definecolor{tcolor}{rgb}{0.1, 0.1, 0.4}
\newcommand{\mytrans}[1]{\textcolor{tcolor}{\textit{(#1)}}}

\begin{table*}[t]
\centering
\scriptsize
\setlength{\tabcolsep}{3.5pt}
\renewcommand{\arraystretch}{1.1}
\begin{tabular}{p{0.12\linewidth} p{0.06\linewidth} p{0.47\linewidth} p{0.25\linewidth}}
\toprule
\textbf{Content} & \textbf{Source} & \textbf{Question} & \textbf{Answer} \\
\midrule
\textbf{(1)} Geography & Certamen &
Which of these was farthest west in the Roman empire: 
A: Tarraconensis \quad B: Cappadocia \quad C: Calabria \quad D: Pannonia &
A \\
\midrule
\textbf{(2)} History & Certamen &
Which of the three women, Cornelia, Pompeia, or Calpurnia, was present during the Bona Dea festival that Clodius Pulcher infiltrated while dressed as a woman?
 &
Pompeia \\
\midrule
\textbf{(3)} Literature & Certamen &
Give the Latin title of the shortest of Plautus’ surviving plays. &
Curculiō \\
\midrule
\textbf{(4)} Mythology & NLE 2015 &
Quis sum? Ego dē Olympō ad terram dēscendō. Sum nūntius deōrum. Ālās in pedibus meīs habeō. \mytrans{Who am I? I descend from Olympus to earth. I am the messenger of the gods. I have wings on my feet.} 
A: Neptūnus \quad B: Mercurius \quad C: Iānus \quad D: Mars &
B \\
\midrule
\textbf{(5)} Vocabulary & \textit{jun-schol} &
What is the feminine equivalent of *gener*? \textit{What is the feminine equivalent of *son-in-law*? } &
nurus \mytrans{daughter-in-law} \\
\midrule
\textbf{(6)} Grammar & Certamen &
Dīc fōrmam plūrālem nominis “sceleris.” 
\mytrans{Say the plural form of the noun “sceleris.”} &
scelerum \\
\midrule
\textbf{(7)} Lit. Devices & NLE 2025 &
What figure of speech is seen in this line from Ennius? Spernitur ōrātor bonus, horridus mīles amātur. 
A: anaphora \quad B: litotes \quad C: polysyndeton \quad D: chiasmus &
D \\
\midrule
\textbf{(8)} Read. Comp. & NLE 2020 &
The tone of Juno’s speech throughout this passage is: 
A: conciliatory \quad B: humble \quad C: persuasive \quad D: condemning &
D \\
\midrule
\textbf{(9)} Scansion & Certamen &
Respondē Latīnē: Dum legis Aeneidem, vidēs haec verba Vergilī: “Conticuēre omnēs intentīque ōra tenēbant.” Quot dactylī sunt in versū?
\mytrans{Respond in Latin: While you read the \emph{Aeneid}, you see this verse by Vergil: “...” How many dactyls are in the verse?} &
duo \\
\midrule
\textbf{(10)} Scansion & \textit{lat-pros} &
What is the name of each foot in the following line of poetry? Give your answer as a comma-separated list. 
Intĕ|gēr vī|tæ, scĕlĕ|rīsquĕ | pūrŭs, &
Trochee, Spondee, Dactyl, Trochee, Trochee \\
\midrule
\textbf{(11)} Scansion & \textit{lat-pros} &
Form the following line into hexameter or pentameter verse by changing the position of one word. The word whose position needs to be changed is marked by the * symbol. 
Ipse dei clypeus terrâ cùm *imâ* tollitur, &
Ipse dei clypeus terrâ cùm tollitur imâ. \\
\midrule
\textbf{(12)} Translation & \textit{jun-schol} &
Put into Latin — “Caius must spare (parco, gerundive) Lucius.” &
Lucio a Caio parcendum est. \\
\midrule
\textbf{(13)} Translation & Certamen &
Now translate: sex ursae in silvā erant. &
“six bears were in the forest”; “there were six bears in the forest”; “there were six female bears in the forest”; “six female bears were in the forest” \\
\bottomrule
\end{tabular}
\caption{Examples of questions across content types, sources, and formats. If the original question or answer is in Latin, we provide an \mytrans{italicized translation}
for the reader, which is not in the actual dataset.
Details described in \S\ref{sec:sources}--\ref{sec:descr}.}
\label{tab:examples_questions}
\vspace{-4mm}
\end{table*}

While many multilingual QA benchmarks have been introduced,
coverage can be limited for low resource languages, and non-existent for historical languages; as far as we know, none of them include Latin.\footnote{The following all exclude Latin: 
\citet{Artetxe:etal:2019, clark-etal-2020-tydi, lewis-etal-2020-mlqa,
longpre-etal-2021-mkqa,
wang2024mmluprorobustchallengingmultitask, bandarkar-etal-2024-belebele};

\citet{xuan2025mmluproxmultilingualbenchmarkadvanced, thellmann2024multilingualllmevaluationeuropean}.} In addition, these benchmarks generally do not include bilingual or mixed-language tasks where both the question and answer can interleave two languages. This gap is particularly relevant for Latin, which has long been taught as a second language in bilingual educational settings, and some researchers may need cross-lingual capabilities; for example, to conduct LLM-based analysis of Latin corpora using English-language instructions or questions.


To address these gaps, we introduce a benchmark dataset for question answering and translation in mixed Latin and English settings, comprising approximately 7,800 question–answer pairs. The questions are derived from a diverse set of pedagogical sources, including standardized exams, quizbowl-style trivia, and textbooks spanning the 1800s to the present. These materials represent centuries of pedagogical practice in teaching Latin, ranging from factual recall of mythology or history to more complex reasoning about syntax, translation, and poetic scansion. The resulting dataset is richly annotated with information about the question's format (multiple choice or short answer), content (10 categories), and language; the answer's language; whether the question requires multiple steps of reasoning ("multihop"); and for translation questions, whether they put constraints on the expected translation.

As a case study, we evaluate three LLMs: two open-source (LLaMa 3 and Qwen QwQ) and one commercial (OpenAI's o3-mini). All exhibit general comprehension but struggle more with skill-oriented questions, such as scansion or translation, which require structured reasoning and linguistic precision. While the reasoning-focused model QwQ shows some advantages on literary and metrical tasks, overall performance is lower than LLaMa's.

Our contributions include:
\begin{itemize}
    \vspace{-1mm}
    \item \textbf{A new mixed-language QA and translation benchmark} for Latin and English, the first of its kind, constructed from education resources spanning multiple centuries.
    \vspace{-1mm}
    \item \textbf{A fine-grained taxonomy of question types} that distinguishes factual knowledge from reasoning skills, enabling more targeted evaluation of LLM capabilities.
    \vspace{-1mm}
    \item \textbf{An evaluation of contemporary open-source LLMs}, revealing gaps in their knowledge and reasoning abilities.
\end{itemize}

\noindent
By making this dataset publicly available, we aim to provide a foundation for evaluating and improving LLMs in specialized linguistic and cultural domains.


\section{Related Work}

Question answering is a staple of Latin learning, though one which recent research suggests the field can benefit from ``insight into the realtime comprehension of Latin'' (\citealp[pg.~298]{bextermoller_2018}; see also \citealp{kuehnast_2024}). Outside the classroom, Latin students have long enjoyed question answering of a different kind, that is ``quiz bowl''-style competitions like Certamen that allow students to ``demonstrate their knowledge of the ancient peoples, languages, and cultures, and the relationships between those topics and the modern world.'' \cite{juniorclassicalleague_2025}. Students can avail themselves of the National Latin Exam for ``opportunity to challenge themselves and measure their growth in the study of the Latin language and Greco-Roman culture.''
\cite{aclnjcl_2024}

Standardized exams are routinely used for benchmarking LLM performance (e.g. SAT, GRE, AP exams in \citealp{openai_2023}); this is the first foray into LLM-assisted QA applied to these Latin pedagogical assessments. While there is a great deal of NLP work in Latin concerning language modeling and tool development \cite{riemenschneider-frank-2023-exploring, bamman_2020}, computational work on pedagogical applications is limited \cite{kuehnast_2024, schulz_2021}; see also \citet{ross_2023}). More general discussions of ML and AI work involving the Latin language (and other low-resource ancient languages) can be found in \citet{sommerschield_2023}. 

Bilingual QA is an active area of research, usually with a narrow focus on a particular language pair \cite{Zhang2021BiRdQAAB, Paschoal2021PirAB, Kalahroodi2025PersianMedQAEL, Mukanova2024DevelopmentOA}. Our dataset adds to this growing body of work.




Parallel sentence data for Latin remain limited, which constrains the development and evaluation of neural machine translation (NMT) systems. Several studies focus on automatic sentence alignment methods to construct parallel sentence datasets \cite{yousef-etal-2022-automatic-translation, Craig2023TestingTL}. 
These efforts are often situated within digital humanities or corpus linguistics research and typically target well-known literary works with existing translations. 


Using either automatically aligned sentences or manually curated alignments, several studies have trained NMT systems for Latin. Prior work includes transformer-based translation systems for Spanish-Latin \cite{martinez-garcia-garcia-tejedor-2020-latin}, English-Latin \cite{rosenthal2023machina}, and German-Latin \cite{fischer-etal-2022-machine}. Because Latin translation is a low-resource task, these systems often incorporate techniques such as transfer learning from higher-resource languages (e.g., Italian) or the integration of morphological information to improve performance \cite{rosenthal2023machina, fischer-etal-2022-machine}. More recent work has also explored the use of generative LLMs for Latin-German translation \cite{volk-etal-2024-llm}.

However, the datasets used in these studies are typically drawn from a narrow set of sources, including the Bible, classical literature, and other widely translated historical texts \cite{martinez-garcia-garcia-tejedor-2020-latin, rosenthal2023machina}. While these corpora are valuable for training NMT systems and for translation studies, their reliance on canonical texts presents challenges for evaluating LLM-based translation systems. Passages from these works and their standard translations are widely circulated in digital corpora, making it likely that they appear frequently in the pretraining data of commercial LLMs. Consequently, evaluations using these datasets may partially reflect memorization rather than a model’s ability to generalize to unseen texts.

To partially mitigate this concern, our evaluation dataset is constructed from pedagogical sources, including exams, trivia, and 19th-century textbooks. Although we cannot guarantee that these materials were excluded from LLM pretraining data, they are less widely circulated than canonical literary translations and therefore less likely to appear repeatedly in training corpora. Our dataset also includes features uncommon in existing Latin translation datasets, such as constrained translation exercises that require the model use specific vocabulary or grammatical constructions, and multiple reference translations.

\section{Data Sources}  \label{sec:sources}
We construct our dataset from four sources, including two textbooks, one set of multiple choice exams, and one set of quizbowl-style trivia questions (Table \ref{tab:data_sources}). When looking for potential sources of data, we aimed for a diversity of question types, both in terms of format and content.

Certamen is a quizbowl-style trivia game played competitively by students studying Latin, Greek, and Classical civilizations \cite{juniorclassicalleague_2025}. Questions cover both language-specific content such as grammar and translation, as well as cultural and historic knowledge. Students can play at three levels of difficulty: novice, intermediate, and advanced.\footnote{\url{https://www.njcl.org/NJCL-Convention/Convention-Contests/Certamen}} 

The National Latin Exam (NLE) is an annual multiple-choice assessment administered to students studying Latin \cite{aclnjcl_2024}.\footnote{\url{https://www.nle.org/}} As of 2025, there are 8 possible exams offered per year, each around 36-40 questions. The exams include beginner, intermediate, and advanced levels, and dedicated exams for reading comprehension of prose and poetry. Like Certamen, the NLE covers both cultural and linguistic knowledge, dedicating half of the exam to grammar and vocabulary and the other half to knowledge of the Roman world. 

Past years' questions from Certamen and NLE are publicly available on their respective websites, but they are copyrighted and not in a structured format ready to be used for NLP applications.

We also sought questions from non-copyrighted historical sources,
by searching through scans of Latin textbooks on HathiTrust\footnote{\url{https://www.hathitrust.org/}} and the Internet Archive\footnote{\url{https://archive.org/}} that had corresponding answer keys. We tried to focus on books which had a variety of question types; many books were excluded because they only contained translation questions.

We settled on two 19th century textbooks. The first, \textit{Latin Grammar and Junior Scholarship Papers} \cite{raven1884latin} contains complex, multipart short answer questions mostly covering vocabulary, grammar, and translation. 
The second, \textit{Exercises in Latin Prosody and Versification} \cite{bradley1823exercises} has challenging questions related to scansion---describing the formal structure of a poetic line according to its long and short syllables.

Certamen and the NLE are based in the United States, while both textbooks were published in England. As a result, the dataset focuses on English-language and Western pedagogy.

\section{Method: Dataset Curation} \label{sec:curation}
During each step of our data curation pipeline, if we used a language model for cleanup or annotation, we performed manual review and intervention of its output.

\paragraph{OCR}
We obtained PDF scans of textbooks and their answer keys from Google Books, and PDFs of the National Latin Exams (NLE) and keys from the NLE website. 

For Certamen, we accessed a publicly available Word document containing questions from 1996–2009, which could be exported directly to plain text, eliminating the need for OCR.

We used Gemini-1.5-pro \cite{geminiteam2024gemini15unlockingmultimodal} to perform PDF text extraction of the NLE texts and OCR of the textbook scans.

A notable challenge involved the breve (˘), a diacritic used to mark short vowels. Although the breve is not commonly used in Latin writing, it appears extensively in \textit{Exercises in Latin Prosody and Versification}, where the distinction between long and short vowels is essential to the content. OCR inconsistently captured this symbol, resulting in large portions of unusable text. 

We manually corrected the accent marks for a small subset of poetry- and scansion-related questions (61 total) from \textit{Exercises in Latin Prosody and Versification}, discarding the rest. Similarly, accent marks were manually corrected in \textit{Latin Grammar and Junior Scholarship Papers} when they were relevant to questions on poetry and scansion.

\paragraph{Alignment of questions to answer keys}
We used a combination of regular expressions and GPT-4o \cite{openai2024gpt4ocard} to align questions with their corresponding answer keys.

First, regular expressions were applied to segment each text into a semi-structured format,
extracting metadata such as chapters, sections, or grouped numbered text blocks, each representing one or several related questions (for example, multipart questions). The same procedure was applied to both question texts and answer keys, producing roughly aligned pairs based on shared structural information such as section headings.

Certamen materials already contained both questions and answers within the same document, so no separate alignment was needed. In this case, regular expressions were used only to identify sections and extract individual questions.

\textit{Exercises in Latin Prosody and Versification} presented the greatest challenge, as it interleaved lessons and explanatory passages with the exercises of interest. We used regular expressions to isolate sections containing exercises and then further subdivided them into individual questions.

For the National Latin Exam (NLE), the formatting was highly consistent, enabling full alignment through regular expressions alone.

For all other sources, we provided Gemini-2.0-Flash with the segmented question and answer text blocks. The model converted these into structured JSON representations, each containing a clearly paired question and answer.

\paragraph{Classify question metadata} 
For each question-answer pair, we used GPT-4o to perform an initial zero-shot classification of several metadata features, then manually reviewed and corrected them.
\begin{itemize}
    \vspace{-1mm}
    \item \textbf{Question format}: multiple choice (MC), or short answer.
    \vspace{-2mm}
    \item \textbf{Question content}: One of 10 possible labels, either knowledge-based (mythology, literature, history, vocabulary, geography) or skill-based (translation, grammar, reading comprehension, scansion, literary devices). This taxonomy is based on lists of topics that Certamen and NLE explicitly intend to cover,\footnote{Certamen and NLE do not have per-question content labels, necessitating zero-shot classification.
    Both aim to cover language and cultural content.
    Specifically, Certamen sources questions on grammar and vocabulary, etymology, mottoes, mythology, politico-military history and geography, material culture and social history, and literature (as per the categories listed on the \href{https://docs.google.com/document/d/1fmB0yTtgaL7oMy-7TV8eghDDeOPIc5XjrqngNXg3EB4/edit?tab=t.0}{Certamen Source List}).
    The \href{https://www.nle.org/what-is-the-national-latin-exam}{NLE website} states it includes questions on ``grammar, comprehension, mythology, derivatives, literature, Roman life, history, geography, oral Latin, and Latin in use in the modern world.''
    (URLs accessed March 2026.)
    } 
    and aligns with broader classifications used in Classics education \cite{Canfarotta_Tosto_Casado-Munoz_2022, Adema+2019+35+59, vereeck2024revered}.
    \vspace{-2mm}
    \item \textbf{Question language and answer language}: each could be either English or Latin. Notably, the question language is the language of the instructions or question, not the primary or majority language present in the question text. This distinction is most clear for translation tasks: the question is the instruction (``put into Latin'' or ``verte in Anglicum'') whereas the source language is the language being translated from. We do not explicitly classify the source language for translation questions, but it can be inferred from the answer language (which is equivalent to the target language).
    \vspace{-2mm}
    \item \textbf{Multihop reasoning}: whether the question requires reasoning through intermediate steps in order to reach the final answer. Multihop reasoning is an ongoing area of focus in NLP research, both for training and evaluation \cite{yang2018hotpotqa, tang-etal-2021-multi, mavi2024multihopquestionanswering, zhang-etal-2024-end}.
    Multihop questions are a common feature of trivia and quizbowl-style datasets in general \cite{rodriguez2021quizbowlcaseincrementalquestion, kabir-etal-2024-make}, so it is unsurprising we found examples in the Certamen dataset. Multihop questions let us test the skill of reasoning LLMs.
    \vspace{-2mm}
    \item \textbf{Constrained translation}: whether the question specifies constraints on the translation, such as requiring a specific lemma or grammatical construction be used.
\end{itemize}

\noindent
Some of these attributes---the format, language, and whether a translation is constrained---are straightforward, but the question content and whether a question requires multihop reasoning are more subjective. For example, history and mythology often overlap with literature. While there may be ambiguity in these cases, we are confident in the broader distinction between knowledge and skill-based questions. Similarly, the definition of a multihop question is vague, as it depends on what is considered a distinct step of reasoning.

A subset of the NLE and Certamen questions also have metadata related to question difficulty. We preserve these difficulty labels in the final dataset, but do not perform automatic classification on questions that did not already have a difficulty label. For this reason, we also do not assess LLMs' performance as it relates to question difficulty.

This metadata enables fine-grained analyses of model performance across different aspects of the dataset.
\vspace{-2mm}

\paragraph{Cleanup and Refinement}
\begin{table}[h!]
\centering
\scriptsize
\setlength{\arrayrulewidth}{0.5pt}
\arrayrulecolor{gray!70}
\begin{tabular}{p{0.08\linewidth} p{0.52\linewidth} p{0.22\linewidth} }
\hline

\textbf{id} & \textbf{question} & \textbf{answer} \\ \hline
\rowcolor{lightpurple!25}
\rowcolor{lightpurple!25}
95.1.2 & Is the gender of the Latin noun ``legio'' determined by form or meaning? & form \\ \hline

\rowcolor{lightblue!25}
95.7.1 & Give the perfect form of ``molo.'' & molui \\ \hline
\rowcolor{lightblue!25}
95.7.2 & Give the supine form of ``molo.'' & molitum \\ \hline
\rowcolor{lightblue!25}
\ldots &  & \\ \hline
\rowcolor{lightblue!25}
95.7.29 & Give the supine form of ``sano.'' & sanatum \\ \hline

\rowcolor{lightpink!25}
95.8.1 & Put into Latin— ``It is all over with the state.'' & Actum est de republica. \\ \hline
\rowcolor{lightpink!25}
\ldots &  & \\ \hline
\rowcolor{lightpink!25}
95.8.4 & Put into Latin— ``Sixteen years after the fall of Carthage.'' & Sexti decimo anno post Carthaginem deletam. \\ \hline
\end{tabular}
\caption{Multipart questions from Figure \ref{fig:pdf_ex} after being broken into standalone questions}
\label{tab:multipart_questions}
\end{table}
We performed a series of cleanup and refinement steps using GPT-4o and manual review to improve data quality and consistency.

First, \textbf{multipart questions} were split into multiple standalone questions, with any references to previous parts disambiguated (Table \ref{tab:multipart_questions}). We then filtered out \textbf{unanswerable or invalid questions}, such as those referencing diagrams or missing context, containing multiple equally valid short answers, or containing OCR errors. Since Certamen is meant to be played as a quizbowl-style competition, some questions may instruct the player to perform an action or to comment on actions performed by the moderator. For example, the moderator points at their eye and asks \textit{Quae pars capitis est haec?} (what part of the head is this?). These types of questions were also filtered out.

Next, for ease of evaluation, we filtered out short answer questions whose answers were longer than one whitespace-delimited word. 

For short answer translation questions, we filtered out questions whose answers were shorter than three whitespace-delimited words. This filtering does not apply to translation questions in MC format. For these short-answer, long-form translation questions, we create \textbf{multiple explicit reference translations} when appropriate. For example, in row 13 of Table \ref{tab:examples_questions}, our final dataset has 4 gold translations made explicit from the original answer in Certamen: "THERE WERE SIX (FEMALE) BEARS IN THE FOREST / SIX (FEMALE) BEARS WERE IN THE FOREST."

\begin{figure}[h!]

    \centering
    \begin{minipage}[t]{0.50\linewidth}
        \centering
        \vspace{0pt} 
        \includegraphics[width=\linewidth]{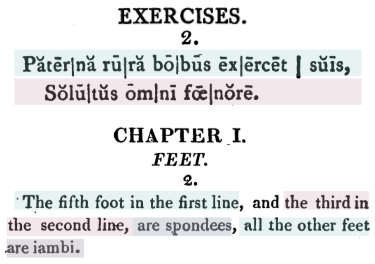}
    \end{minipage}
    \hfill
    \raisebox{-0.01\textheight}{
    \begin{minipage}[t]{0.46\linewidth}
        \centering
        \vspace{0pt} 
        \scriptsize
        \begin{tabular}{@{}l p{0.6\linewidth}@{}} 
            \toprule
            \textbf{id} & \textbf{answer} \\
            \midrule
            \rowcolor{lightblue!25}
            app.I.2.1 & iambus, iambus, iambus, iambus, spondee, iambus \\ \hline
            \rowcolor{lightpink!25}
            app.I.2.2 & iambus, iambus, spondee, iambus \\
            \bottomrule
        \end{tabular}
    \end{minipage}
    }

    \caption{\textbf{(left)} Original question and answer from \textit{Exercises in Latin Prosody and Versification}, and \textbf{(right)} Answers to the questions, reworded for ease of evaluation.}
    \label{fig:pdf_scansion_examples}
    \vspace{-2mm}
\end{figure}

We also simplified verbose answers for specific question types, particularly prosody and scansion exercises, to make evaluation more reliable (Figure \ref{fig:pdf_scansion_examples}).

Some Certamen short-answer questions explicitly gave a candidate list of single-word options, so we found it more natural to reformat them as multiple choice (e.g., "Which of the following Latin nouns does not belong because of gender: cor, agricola, senātus, pēs, leō").
Certamen has no original MC questions, so all 317 MC questions from Certamen were converted from their original SA format using regex. 

Finally, we duplicated the scansion-related questions from \textit{Exercises in Latin Prosody and Versification}
by translating their English instructions into Latin, ensuring balanced representation across both languages and allowing us to precisely examine the effect of the question language on LLM performance.

\section{Dataset Description}  \label{sec:descr}
\begin{table}[h!]
\centering
\small
\begin{tabularx}{\linewidth}{l | r r r | r}
\toprule
\textbf{Source} & \textbf{MC} & \textbf{1-W SA} & \textbf{Long A.} & \textbf{Total} \\
\midrule
Certamen & 317 & 4540 & 970 & 5827 \\
NLE & 855 & 0 & 0 & 855 \\
\textit{Lat-Pros} & 0 & 0 & 122 & 122 \\
\textit{Jun-Schol} & 0 & 675 & 350 & 1025 \\
\midrule
\textbf{Total} & 1172 & 5215 & 1442 & \textbf{7829} \\
\bottomrule
\end{tabularx}
\caption{Source of data verseus question formats (MC=multiple choice; 1-W SA=one-word short answer; Long A.=long answer).}
\label{tab:stats_format_by_source}
\end{table}
Our final dataset consists of 7,829 question-answer pairs, with the most common format being 1-word short answer (SA) sourced from Certamen (Table \ref{tab:stats_format_by_source}). Multiple choice questions have between 3-7 options. Both MC and 1-word SA can be evaluated with accuracy, but long answers include a variety of output types which require separate evaluation strategies (see \S\ref{sec:eval_metrics}).

\begin{table}[h!]
\setlength{\tabcolsep}{3pt}
\centering
\small
\begin{tabular}{l | rrrr | r}
\toprule
& \multicolumn{4}{c}{\textbf{Question-Answer Lang}} & \\
\textbf{Content} & \textbf{En--En} & \textbf{La--En} & \textbf{En--La} & \textbf{La--La} & \textbf{Total} \\
\midrule
  Geography & 73 & 1 & 107 & 1 & 182 \\
  History & 253 & 0 & 685 & 3 & 941 \\
  Literature & 85 & 2 & 311 & 0 & 398 \\
  Mythology & 230 & 1 & 1283 & 8 & 1522 \\
  Vocabulary & 698 & 9 & 733 & 21 & 1461 \\
\midrule
  Grammar & 192 & 122 & 894 & 106 & 1314 \\
  Lit. Devices & 28 & 1 & 27 & 0 & 56 \\
  Read. Comp. & 352 & 7 & 26 & 0 & 385 \\
  Scansion & 21 & 20 & 59 & 42 & 142 \\
  Translation & 854 & 91 & 483 & 0 & 1428 \\
\midrule
\textbf{Total} & 2786 & 254 & 4610 & 179 & \textbf{7829} \\
\bottomrule
\end{tabular}
\caption{Counts of question--answer language pairs versus question content type. Top rows are \textbf{knowledge-based} and bottom rows are \textbf{skill-based}.}
\label{tab:lang_by_content}
\end{table}
Table \ref{tab:lang_by_content} shows the number of QA pairs by language and content type. English questions with Latin answers make up the majority (4610), followed by English questions with English answers (2786). The amount of questions asked in Latin is much smaller, with only 433 total. English questions are also spread nicely across content types, but Latin questions are more sparse. 

\begin{table}[h!]
\centering
\small
\begin{tabular}{l | r r | r}
\toprule
\textbf{Type} & \textbf{La→En} & \textbf{En→La} & \textbf{Total} \\
\midrule
Unconstrained & 818 & 285 & 1103 \\
Constrained & 17 & 146 & 163 \\
\midrule
\textbf{Total} & 835 & 431 & \textbf{1266} \\
\bottomrule
\end{tabular}
\caption{Translation direction (src→target) by constraint type for long-form (3+ word) translation questions.}
\label{tab:stats_translation_by_type}
\vspace{-2mm}
\end{table}
For translation questions specifically, Table \ref{tab:stats_translation_by_type} shows there is an over-representation of Latin→English, but the reverse direction still has a sizable amount of examples.
For Latin as the target, there are an average of 2.2 reference translations, max of 30; for English, an average of 2.3, max 48.
About 13\% (166) of the translation questions are constrained, with most of those being English→Latin (146).

\begin{table}[h!]
\centering
\small
\begin{tabular}{l | r r | r}
\toprule
\textbf{Content} & \textbf{Regular} & \textbf{Multihop} & \textbf{Total} \\
\midrule
Geography & 99 & 57 & 156 \\
History & 612 & 280 & 892 \\
Literature & 303 & 79 & 382 \\
Mythology & 1181 & 303 & 1484 \\
Vocabulary & 968 & 266 & 1234 \\
\midrule 
Grammar & 925 & 77 & 1002 \\
Lit. Devices & 44 & 2 & 46 \\
Translation & 52 & 4 & 56 \\
Scansion & 17 & 2 & 19 \\
\midrule
\textbf{Total} & 4201 & 1070 & \textbf{5271} \\
\bottomrule
\end{tabular}
\caption{Counts of regular vs.\ multi-hop 1-word SA questions and their question content.}
\label{tab:multihop_by_content}
\end{table}
Finally, we examine the 1-word short answer questions. About 20\% (1070) are multihop (Table \ref{tab:multihop_by_content}).
The skill-based questions have lower proportions of multihop questions compared to the knowledge-based questions. In particular, geography and history have the highest proportion of multihop questions (37\% and 31\%, respectively).

We provide examples of questions across all language pairs, content types, formats, and sources in Table \ref{tab:examples_questions}.

\section{Experiments}
To illustrate the utility of our dataset to benchmark LLMs,  we propose with a set of prompts and evaluation metrics, applied to three current LLMs.

\subsection{Experimental Setup}
\paragraph{Models} We evaluate two open-source LLMs---LLaMa 3.3 \cite{grattafiori2024llama3herdmodels} and Qwen QwQ \cite{qwen2.5,qwq32b}---and one commercial model, OpenAI's o3-mini.\footnote{\url{https://openai.com/index/introducing-o3-and-o4-mini/}}\footnote{Version identifiers: \texttt{meta-llama/Llama-3.3-70B-Instruct-Turbo}, \texttt{Qwen/QwQ-32B}, and \texttt{o3-mini-2025-01-31}. Open models were accessed with the \href{https://www.together.ai/}{Together AI API}.} 
LLaMa3 is a 70 billion parameter instruction tuned model with strong multilingual performance. Qwen QwQ is a smaller, 32 billion parameter model trained from Qwen 2.5 using reinforcement learning (RL) with verifiable outcome-based rewards and a standard reward model. 
In theory, good reasoning ability could be applicable to our skill-based questions that involve more problem-solving. 

\paragraph{Prompts} 
We always provide the system prompt \textit{You are a Classicist with expert knowledge in Greek and Roman history, language, and culture.}

For MC questions, we instruct the model to end their response with the letter of the correct answer. Similarly, for 1-word SA, we ask it to end its response with a single word as its answer.

When prompting the two open models, we use a temperature of 0.6 and top-$p$ of 0.95, the recommended settings for QwQ. These parameters were not tunable for o3-mini using the OpenAI API.

\paragraph{Evaluation Metrics} \label{sec:eval_metrics}
For multiple choice questions, we evaluate the accuracy of the predicted letter choice.

For 1-word short answer questions, we use exact match (EM) accuracy, after normalization (lowercasing, stripping punctuation and whitespace, JV replacement,\footnote{In classical Latin orthography, both I and J are represented by a single letter; same with U and V. It is a common Latin NLP preprocessing step to normalize words to collapse these letter distinctions. For example, the Classical Language Toolkit, an NLP pipeline for pre-modern languages \cite{johnson-etal-2021-classical}, includes a \href{https://v0.cltk.org/en/latest/latin.html}{JV replacer} and recommends using it during preprocessing.} 
normalizing macrons, accents, and ligatures).
For the majority of questions, normalizing accents will not affect the correctness of the answer. Only questions that ask for vowel quantity to be marked are affected, which is less than 20 questions in the dataset.

For a subset of scansion questions ("feet identification"), we use mean per-item accuracy (see row 10 of Table \ref{tab:examples_questions} for an example). Each question gives one line of poetry and asks for the name of each metrical foot in the order it appears in the verse. For a single line, the number of feet typically ranges from 3-6, and partial correctness is allowed.

For another subset of scansion questions ("meter manipulation"), we use accuracy (see row 11 of Table \ref{tab:examples_questions}).
Each question gives a line of poetry and asks that the position of a single word be changed in order to make the verse valid pentameter or hexameter. Since only one word should be moved, we consider it correct (1) if the word is in the correct position and incorrect (0) otherwise.

Finally, for long-form translation questions, we report the BLEU score \cite{papineni-etal-2002-bleu}.\footnote{We use the sacreBLEU implementation \cite{post-2018-call}.}
Although BLEU has received criticism for its low correlation with human judgments \cite{callison-burch-etal-2006-evaluating, karpinska-etal-2022-demetr}, the quality of newer, fine-tuned neural metrics for Latin is untested, and there is a lack of gold parallel sentence data available for developing such methods.

\subsection{Results}
\paragraph{MC and 1-Word SA Accuracy}
Overall accuracy (excluding long-form translation), is 71.92\% for LLaMa3, 68.11\% for QwQ, and 68.61\% for o3-mini. Performance is much higher on MC formatted questions, with LLaMa at 90.25\%, QwQ 90.86\%, and o3-mini 91.80\%. 

All models generally perform better on the knowledge-based questions than the skill-based questions, and LLaMa performs best overall. LLaMa has 74.27\% accuracy for knowledge questions, 64.77\% for skill; QwQ 71.08\% knowledge and 61.63\% skill; and o3-mini 69.46\% knowledge and 66.76\% skill. The gap in performance between knowledge and skill questions is noticeably smaller for o3-mini.

\begin{figure}[h!]
    \centering
    \includegraphics[width=\linewidth]{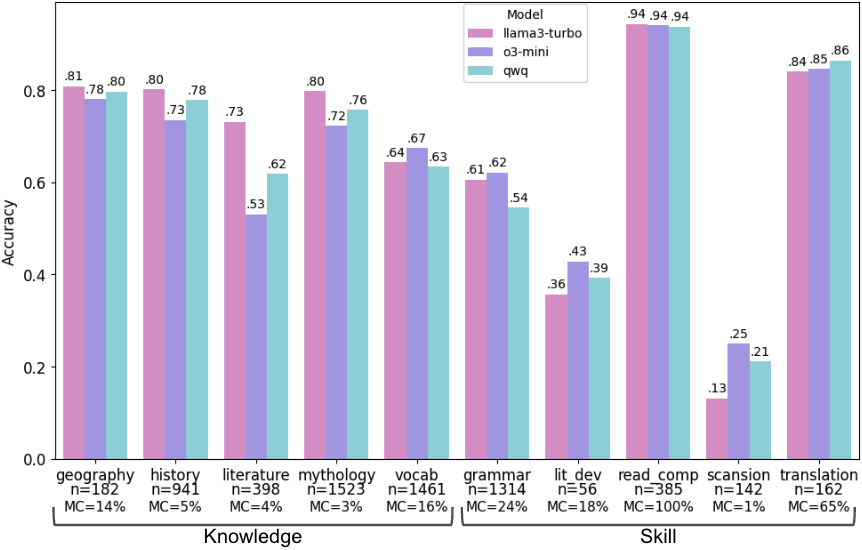}
    \caption{Accuracy by question content. Knowledge categories are on the left and skill categories on the right. Includes both MC and 1-word SA questions.}
    \label{fig:res_acc_content}
\end{figure}
In Figure \ref{fig:res_acc_content}, LLaMa is the best performer for most knowledge-based content types, and o3-mini is worst performing. The reasoning models, QwQ and o3-mini, show an advantage over LLaMa on literary devices and scansion questions, although the best scansion accuracy (25\%, o3-mini) is still far behind the other question types.

High accuracy on the skill-based reading comprehension and translation questions is likely due to their questions being majority MC format.

\begin{figure}[h!]
    \centering
    \includegraphics[width=\linewidth]{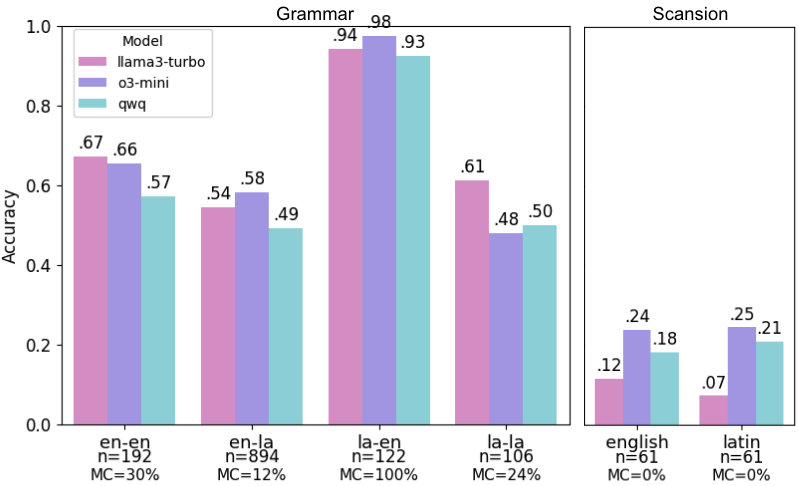}
    \caption{Accuracy by question-answer language pair, for Grammar questions (left), and by question language for Scansion questions (right).}
    \label{fig:res_acc_lang}
    \vspace{-2mm}
\end{figure}
\paragraph{Effect of Question Language} To analyze the effect of the question language on performance, we only report accuracy on Grammar and Scansion questions, since they have enough examples for each language pair. 

Although the Latin-English Grammar pairs have the highest accuracy in Figure \ref{fig:res_acc_lang}, these are entirely MC formatted questions from NLE.

Keeping the answer language fixed to Latin, LLaMA and QwQ have better performance when a Grammar question is asked in Latin rather than in English. This gap is larger for LLaMa (61\% La-La, 54\% En-La) than for QwQ (50\% La-La, 49\% En-La). For o3-mini, this effect is reversed, with a 10\% drop in accuracy.

Similarly, QwQ performs better on scansion questions asked in Latin (21\%) than those asked in English (18\%). However, this is reversed for LLaMa, with 7\% accuracy on scansion questions asked in Latin and 12\% on those asked in English. Accuracy is about the same for o3-mini for English (24\%) and Latin (25\%) questions.

\begin{figure}[h!]
    \centering
    \includegraphics[width=\linewidth]{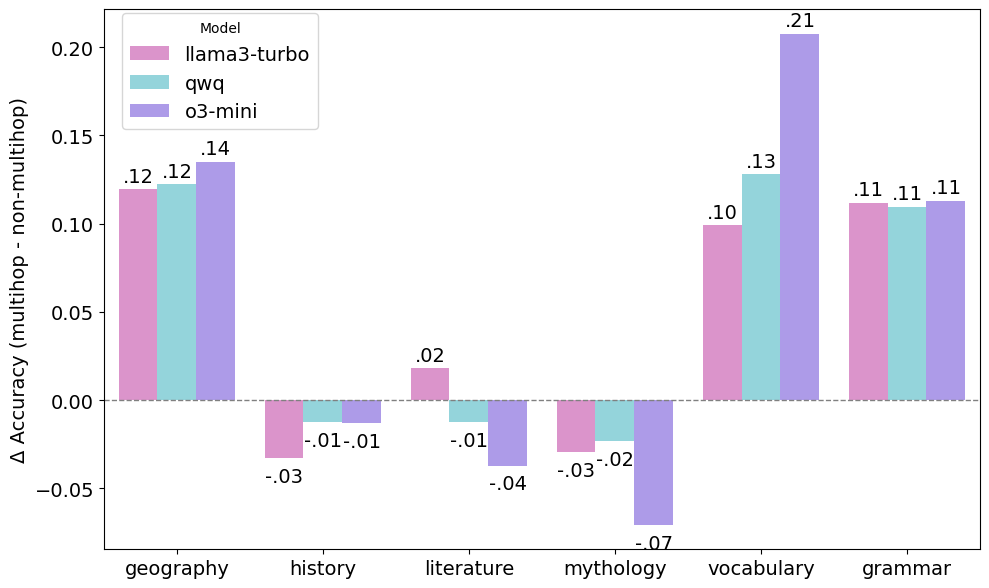}
    \caption{Difference in accuracy: multihop - regular for 1-word SA questions.}
    \label{fig:res_multihop}
    \vspace{-4mm}
\end{figure}
\paragraph{Multihop questions} Surprisingly, all the LLMs performed better overall on the multihop questions (Figure \ref{fig:res_multihop}). Some content types (history, literature, mythology) showed mild regressions on multihop questions. These mixed results may be due to the subjectivity of what qualifies as a distinct step of reasoning. In addition, tasks that require multiple steps of reasoning for a human may have answers that are directly stated in LLM training data. For example, a grammar question may ask for a word to be inflected to a particular form. A human may need to think through multiple steps to come to the correct answer (what declension/conjugation is the word, what are the correct endings for that declension/conjugation, what is the particular ending the question is asking for, and how does that ending combine with this lemma?), but an LLM may have seen all inflected forms neatly formatted together in its training data.

\begin{table}[h!]
\centering
\small
\setlength{\tabcolsep}{3pt}
\begin{tabular}{l r r | r r | r r}
\toprule
 & \multicolumn{2}{c}{\textbf{LLaMA~3}} & \multicolumn{2}{c}{\textbf{QwQ}} & \multicolumn{2}{c}{\textbf{o3-mini}} \\
\cmidrule(lr){2-3} \cmidrule(lr){4-5} \cmidrule(lr){6-7}
\textbf{Setting} & \textbf{Lat} & \textbf{Eng} & \textbf{Lat} & \textbf{Eng} & \textbf{Lat} & \textbf{Eng} \\
\midrule
Unconst. & 25.50 & 45.41 & 18.53 & 39.91 & 23.42 & 43.29 \\
Const.   & 20.88 & 37.46 & 21.52 & 17.95 & 27.25 & 34.68 \\
Overall       & 23.71 & \textbf{45.25} & 19.53 & 39.27 & \textbf{24.67} & 43.14 \\
\bottomrule
\end{tabular}
\caption{BLEU scores for each translation setting and target language. }
\label{tab:res_bleu_scores}
\vspace{-2mm}
\end{table}
\paragraph{Long-form Translation}
Overall, LLaMa is the best model for translating into English, and o3-mini is best for translating into Latin.

There are very few constrained translation questions where English is the target language, so results are less valid.

LLaMa performs worse on constrained translations across each target language setting. QwQ sees a large drop on constrained translation when English is the target, but we observed this was caused by the model not following directions and leaving explanations in its final answer.

QwQ and o3-mini perform better on the constrained Latin translations than the unconstrained ones, outperforming LLaMa. If a model is correctly able to follow the given constraint, then the space of possible translations is smaller, so it should be easier to provide a translation closer to the reference(s).

\section{Discussion and Future Work}
Reasoning abilities are beneficial for some skill-based tasks (scansion, literary devices) but are unable to compensate for poorer foundational knowledge. Considering the added computational cost, it is unnecessary to use reasoning models for most tasks we tested. We also observed QwQ's reasoning ability sometimes prevented it from coming to an answer at all, getting stuck in reasoning loops. However, o3-mini, the other reasoning model, did not have the same issue.

In this paper, we use basic prompting strategies, but more
targeted techniques may need to be developed to improve performance in certain areas such as vocabulary, grammar, and especially scansion.

All models lag behind when translating into Latin versus English. More work should investigate this gap, as well as the effect of the instruction language and the type of constraint present in the instruction.

Future work to flesh out sparser content types and language pairs in the dataset would be especially valuable. Additionally, questions with long, phrase- or sentence-length answers could be added, and automatic evaluation methods could be tested for these questions.

Although we use this dataset for evaluation only, it could also be used for training of MT systems or instruction tuning of larger generative models.

\section{Conclusion}
We present the first benchmark for QA and translation in mixed Latin–English settings, built from over 7000 questions spanning two centuries of pedagogical materials and capturing a wide spectrum of linguistic and reasoning challenges. Our evaluation of three large language models reveals that even strong general-purpose models struggle with skill-based and linguistically precise tasks. We hope this resource will support future research on multilingual and historical language understanding, and serve as a blueprint for building comparable resources in other low-resource settings.

\section{Ethics}
Our dataset is derived from publicly available materials, but some subsets are copyrighted and have distinct terms of use and access.

At the time of writing, we do not plan to redistribute the portions of our dataset sourced from Certamen. The Junior Classical League (JCL) has agreed to host the Certamen portion of our dataset on its website along with their archived Certamen questions.

We will host the subset of our data sourced from NLE. The ACL/JCL National Latin Exam does not allow these materials to be used for generation of profit.

The most up-to-date access to the dataset and details on terms of use will be maintained at:
\href{https://github.com/slanglab/RespondeoQA}{https://github.com/slanglab/RespondeoQA}.

\section{Limitations}
It is possible that our questions exist in LLM pretraining data. However, the performance of the tested models still has room for improvement. Even if our data was seen by the models during training, it is also unlikely to have seen answers aligned to the questions.

Some combinations of question types, content, and languages are sparsely represented in our dataset, so a robust evaluation of performance is not yet possible. We try to limit evaluations to categories that have enough examples.

\section{Acknowledgments}
We would like to thank the UMass NLP group for their feedback and commentary on this project. 
This material is based in part upon work
supported by National Science Foundation award
1845576 (CAREER). Any opinions, findings and conclusions
or recommendations expressed in this material are
those of the authors and do not necessarily reflect
the views of the National Science Foundation.

\section{Bibliographical References}\label{sec:reference}

\bibliographystyle{lrec2026-natbib}
\bibliography{lrec2026-example}

\bibliographystylelanguageresource{lrec2026-natbib}
\bibliographylanguageresource{languageresource}

\end{document}